\documentclass[conference]{IEEEtran}
                                            \IEEEoverridecommandlockouts

                                            \usepackage{amsmath,amssymb,amsfonts}
                                            \usepackage{cite}
                                            \usepackage{graphicx}
                                            \usepackage{booktabs}
                                            \usepackage{multirow}
                                            \usepackage{array}
                                            \usepackage{float}
                                            \usepackage{url}
                                            \usepackage{hyperref}
                                            \usepackage{algorithm}
                                            \usepackage{algorithmic}
                                            \usepackage{xcolor}

\begin{document}

                                            \title{ChainLearn: A Blockchain-Based Capacity-Aware Framework for Federated Ensemble Learning
                                            \thanks{Code is available at: \url{https://github.com/EdddTri/blockchain_capacity}}}

                                            \author{
                                                \IEEEauthorblockN{Karan Sharma}
                                                \IEEEauthorblockA{DA-IICT \\ Gandhinagar, India \\ 202318018@dau.ac.in}
                                                \and
                                                \IEEEauthorblockN{Aditya Tripathi}
                                                \IEEEauthorblockA{DA-IICT \\ Gandhinagar, India \\ 202318046@dau.ac.in}
                                                \and
                                                \IEEEauthorblockN{Rahul Mishra}
                                                \IEEEauthorblockA{IIT Patna \\ Patna, India \\ rahul\_mishra@iitp.ac.in}
                                                \and
                                                \IEEEauthorblockN{Tapas Kumar Maiti}
                                                \IEEEauthorblockA{DA-IICT \\ Gandhinagar, India \\ tapas\_kumar@dau.ac.in}
                                            }

                                            \maketitle

                                            \begin{abstract}

                                            Federated learning is used in medical imaging where privacy prohibits centralizing data. Standard federated algorithms assume homogeneous hardware, identical architectures, and centralized aggregation, which fails when hospitals have unequal compute resources. We propose capacity-aware coordination: measure each hospital's throughput, assign capacity-appropriate architectures (MobileNetV3-Small, EfficientNet-B0, ResNet-50), and combine predictions via weighted ensemble. Weak and strong hospitals can participate without forcing uniform architectures.

                                            We separate on-chain policy from off-chain learning. A Solidity contract stores hospital registration, benchmark hashes, metrics, and weights. Hospitals train locally and submit only hashes and scalars (not parameters). Weighted ensemble inference is computed off-chain.

                                            Experiments on PneumoniaMNIST and DermaMNIST (5 seeds, 3 non-IID levels) show our method achieves lower or equal calibration error versus equal-weight ensemble and competitive accuracy versus FedAvg, FedProx, and FedMD. Communication overhead is 224 bytes per round, a reduction of over 912,000$\times$ compared to FedAvg.

                                            \end{abstract}


                                            \section{Introduction}

                                            Federated learning enables collaborative machine learning in domains where raw data cannot be centrally pooled. Healthcare systems enforce this strongly: patient data is protected by regulation and institutional policy. Hospitals have locally valuable datasets but cannot share patient images directly, so they train local models and coordinate through aggregation.

                                            The majority of practical federated learning systems remain built around an implicit homogeneity assumption: all participants train identical architectures under approximately comparable computational conditions. In canonical parameter-averaging frameworks such as FedAvg, local model tensors must share identical dimensional structure for aggregation to remain valid. This assumption is difficult to satisfy in realistic hospital networks. A tertiary medical center may possess server-grade GPU infrastructure capable of sustaining deep residual architectures, while smaller institutions may operate under CPU-only constraints or lightweight edge environments. Under such conditions, architecture uniformity either excludes weaker participants or underutilizes stronger ones.

                                            This mismatch creates a structural limitation in conventional federated optimization. If all participants must train the same network, the chosen architecture is determined by the weakest participant. If stronger institutions independently train larger models, standard parameter averaging becomes mathematically invalid because tensor dimensions no longer align. Heterogeneous environments force a tradeoff: include weak hospitals (limit model size) or use strong models (exclude weak hospitals).

                                            We solve this by treating architecture as a coordination variable. We benchmark each hospital's throughput, assign it to a Weak/Medium/Strong tier, then assign MobileNetV3-Small, EfficientNet-B0, or ResNet-50 accordingly. Since parameter spaces differ, we aggregate in prediction space: a weighted softmax ensemble replaces parameter averaging.

                                            A second challenge is coordination trust. Standard federated learning uses a central aggregator, which is opaque and represents a single point of failure. We address this with a blockchain coordination layer. A Solidity contract stores registration, signed benchmarks, declared capacity, metrics, and weights. Models remain off-chain. Only hashes and scalars are submitted on-chain.

                                            The blockchain stores policy and metadata only. It does not train models or run inference. Those operations are performed off-chain. Hospitals submit model hashes, confidence estimates, expected calibration error, and architecture identifiers. The contract applies a fixed-point arithmetic weighting rule combining capacity multiplier, calibration-adjusted confidence, and participation history. Final ensemble inference occurs off-chain, where local model outputs are combined through weighted probability averaging.

                                            The contributions of this paper are threefold:

                                            \begin{enumerate}
                                            \item We introduce a capacity-aware heterogeneous federated coordination workflow in which participants are assigned distinct model architectures according to measured local compute throughput, allowing institutions with unequal hardware resources to jointly participate without parameter-space compatibility requirements.

                                            \item We design a blockchain coordination layer that stores identity-bound submissions, enforces architecture-class consistency, and deterministically computes aggregation weights from declared capacity, self-reported confidence, calibration error, and participation history while retaining all model parameters off-chain.

                                            \item We implement a communication-minimal off-chain weighted ensemble strategy that aggregates heterogeneous model outputs in probability space and evaluate it against multiple federated baselines under non-IID medical image classification settings.

                                            \end{enumerate}

                                            \section{Related Work}

                                            \subsection{Classical Federated Learning}

                                            Federated learning emerged as a communication-efficient alternative to centralized distributed optimization by enabling local gradient computation while preserving data locality. The most widely adopted baseline remains FedAvg, in which local participants perform several gradient steps and transmit parameter tensors to a central aggregator for weighted averaging. Under homogeneous architectures and comparable data distributions, this approach remains highly effective.

                                            However, FedAvg depends fundamentally on parameter compatibility:

                                            \begin{equation}
                                            \theta^{global} = \sum_{i=1}^{N} \alpha_i \theta_i
                                            \end{equation}

                                            where all $\theta_i$ must share identical tensor structure. This requirement becomes restrictive when clients differ in hardware capacity or model family selection.

                                            \subsection{Federated Learning Under Heterogeneity}

                                            Several extensions attempt to relax FedAvg assumptions under statistical or systems heterogeneity. FedProx introduces proximal regularization to stabilize local optimization under unequal participant behavior. This improves convergence when local objectives differ but still assumes compatible architectures.

                                            FedMD uses probability-space coordination via distillation. The present system uses blockchain-defined weighting rather than distillation, and explicitly externalizes coordination policy into a smart contract rather than relying on centralized orchestration.

                                            \subsection{Blockchain-Based Federated Coordination}

                                            Blockchain has been repeatedly proposed as a trust-reduction mechanism for federated learning, particularly for participant registration and immutable logging. Many prior systems, however, overextend blockchain responsibilities by implying on-chain model coordination at scales incompatible with realistic neural parameter sizes. Large convolutional backbones cannot be economically stored on-chain, and direct parameter exchange through smart contracts is computationally infeasible.

                                            We minimize on-chain data to: benchmark hash, model hash, scalar confidence, scalar calibration error, and architecture identifier. This reduces chain usage to deterministic policy storage rather than model transport.

                                            \subsection{Federated Learning for Medical Imaging}

                                            Medical imaging remains one of the most compelling domains for federated learning because privacy barriers strongly prohibit centralized pooling of patient scans, yet hospital networks also exhibit strong infrastructure asymmetry. Large tertiary hospitals frequently possess significantly stronger compute than smaller institutions. This asymmetry is precisely what homogeneous FL does not handle. We use medical imaging as a realistic heterogeneous scenario.

                                            \section{Problem Formulation}

                                            Let:

                                            \begin{equation}
                                            \mathcal{H} = \{H_1, H_2, ..., H_N\}
                                            \end{equation}

                                            denote a set of participating hospitals.

                                            Each hospital $H_i$ possesses local private dataset:

                                            \begin{equation}
                                            D_i = \{(x_j,y_j)\}_{j=1}^{n_i}
                                            \end{equation}

                                            and local compute capacity:

                                            \begin{equation}
                                            c_i \in \{weak, medium, strong\}
                                            \end{equation}

                                            determined through benchmark throughput.

                                            A deterministic architecture mapping assigns:

                                            \begin{equation}
                                            A_i = f(c_i)
                                            \end{equation}

                                            such that:

                                            \begin{equation}
                                            weak \rightarrow MobileNetV3Small, \quad medium \rightarrow EfficientNetB0, \quad strong \rightarrow ResNet50
                                            \end{equation}

                                            Each hospital trains local model:

                                            \begin{equation}
                                            M_i = Train(A_i,D_i)
                                            \end{equation}

                                            and computes reliability tuple:

                                            \begin{equation}
                                            R_i = (C_i,E_i,r_i)
                                            \end{equation}

                                            where $C_i$ is average confidence, $E_i$ is expected calibration error, and $r_i$ is rounds participated.

                                            The blockchain coordination function computes:

                                            \begin{equation}
                                            W_i = g(c_i,C_i,E_i,r_i)
                                            \end{equation}

                                            using fixed-point arithmetic.

                                            Because local parameter spaces differ ($\theta_i \not\equiv \theta_j$ when $A_i \neq A_j$), aggregation occurs in probability space:

                                            \begin{equation}
                                            P(x)=\frac{\sum_{i=1}^{N}W_iP_i(x)}{\sum_{i=1}^{N}W_i}
                                            \end{equation}

                                            where $P_i(x)$ is the local softmax output.

                                            The objective is deterministic coordination across heterogeneous architectures $A_i$ subject to the constraints that local data and model parameters do not leave each hospital.

                                            \section{System Architecture}

                                            The proposed framework is organized as a three-layer execution architecture aligned with trust boundaries and runtime responsibilities. Rather than treating blockchain as a computational substrate for machine learning, the design deliberately isolates heavy numerical learning from policy coordination. Neural model execution is computationally unsuitable for smart contract environments, while coordination semantics benefit from deterministic externalization.

                                            The complete architecture consists of:

                                            \begin{enumerate}
                                            \item an on-chain coordination and policy layer,
                                            \item an off-chain hospital execution layer,
                                            \item an experimental orchestration layer used for research evaluation.
                                            \end{enumerate}

                                            These layers correspond directly to the repository directories \texttt{smart\_contracts/}, \texttt{hospital\_node/}, and \texttt{simulation/}, reflecting distinct operational trust domains.

                                            \subsection{On-Chain Coordination Layer}

                                            The blockchain layer is implemented through the Solidity contract \texttt{FLCoordinator}. Its purpose is not to store models or execute inference, but to encode deterministic participation policy and persistent round metadata.

                                            The contract maintains four primary categories of state: hospital registration metadata, per-round submissions, round submitter lists, and ensemble records.

                                            A hospital entry stores:

                                            \begin{equation}
                                            H_i=(addr_i,name_i,c_i,registered_i,h_i,C_i,E_i,r_i)
                                            \end{equation}

                                            where $addr_i$ denotes Ethereum identity, $c_i$ denotes declared capacity class, $h_i$ denotes benchmark hash, $C_i$ denotes latest confidence, $E_i$ denotes latest calibration error, and $r_i$ denotes rounds participated.

                                            \subsection{Off-Chain Hospital Layer}

                                            Each hospital executes all computationally expensive operations locally: benchmark generation, model training, reliability computation, model hashing, and blockchain transaction signing.

                                            This logic resides primarily in \texttt{capacity\_manager.py}, \texttt{blockchain\_client.py}, and \texttt{contract\_integration.py}. The hospital layer performs training and inference. The blockchain layer receives only scalar summaries.

                                            \subsection{Experiment Layer}

                                            The research evaluation layer introduces controlled benchmarking independent of live contract execution. It includes baseline training, non-IID partition generation, ablation analysis, communication estimates, gas estimates, and adversarial simulation, implemented across three scripts:

                                            \begin{itemize}
                                            \item \texttt{experiment.py} --- comprehensive evaluation: seven baselines, five ablations, three non-IID levels, five seeds, adversarial scenarios, and communication and gas cost analysis. The contract weight logic is mirrored in Python for efficient iteration over training loops.
                                            \item \texttt{run\_simulation.py} --- end-to-end blockchain protocol demonstration: starts a local Hardhat node, deploys \texttt{FLCoordinator}, registers hospitals with ECDSA-signed PoC hashes, runs multi-round training, submits updates on-chain, reads contract-computed weights, and records ensemble prediction hashes on-chain.
                                            \item \texttt{simulate\_federation.py} --- lightweight FedAvg protocol demonstration on synthetic data partitions, serving as a standalone protocol sanity check.
                                            \end{itemize}

                                            The contract and Python evaluation use identical arithmetic, enabling accurate testing of policy semantics.

                                            \section{Smart Contract Design and Coordination Semantics}

                                            \subsection{Contract State Model}

                                            The Solidity implementation defines three principal structs:

                                            \subsubsection{Hospital}

                                            \begin{multline}
                                            Hospital=(address,\ name,\ capacityClass,\\
                                            \quad isRegistered,\ pocHash,\ confidence,\ ece,\ rounds)
                                            \end{multline}

                                            This struct stores both registration metadata and evolving reliability state.

                                            \subsubsection{Submission}

                                            \begin{multline}
                                            Submission=(modelHash,\ confidence,\\
                                            \quad ece,\ timestamp,\ modelType)
                                            \end{multline}

                                            A submission is round-scoped.

                                            \subsubsection{EnsembleRecord}

                                            \begin{multline}
                                            EnsembleRecord=(predictionHash,\\
                                            \quad participantCount,\ timestamp)
                                            \end{multline}

                                            The contract stores opaque ensemble evidence rather than prediction tensors.

                                            \subsection{Round State Machine}

                                            The contract operates as a privileged owner-controlled round machine with the lifecycle: registration, round start, submission collection, weight query, and ensemble recording.

                                            The owner invokes $startNewRound()$. Hospitals invoke $submitUpdate()$. The aggregator records $recordEnsemblePrediction()$. Round submissions are decentralized while round initiation is owner-managed.

                                            \subsection{Submission Constraints}

                                            The contract enforces three participation constraints.

                                            \subsubsection{Registered Identity Constraint}

                                            Only registered hospitals may submit ($registered_i = true$).

                                            \subsubsection{Single Submission Constraint}

                                            Each hospital may submit at most once per round.

                                            \subsubsection{Model-Type Constraint}

                                            The declared model type must match the assigned capacity class:

                                            \begin{equation}
                                            modelType_i = f(capacity_i)
                                            \end{equation}

                                            Architecture compliance is enforced at submission time. The Solidity contract rejects mismatches.

                                            \subsection{Fixed-Point Weight Computation}

                                            The weight function is implemented entirely in integer arithmetic. Contract constants are $SCALE = 10000$ and $MAX = 15000$. Capacity multipliers are $weak = 8000$, $medium = 10000$, $strong = 12000$.

                                            The deterministic weight is:

                                            \begin{equation}
                                            W_i = \min\left(\frac{M_i C_i (\text{SCALE}-E_i)}{\text{SCALE}^2} + B_i,\ \ \text{MAX}\right)
                                            \end{equation}

                                            where $\text{MAX}=15000$ prevents any single participant from dominating the ensemble when its reliability/capability product saturates. This cap is enforced at the Solidity level (\texttt{MAX\_WEIGHT} constant in \texttt{FLCoordinator.sol}), ensuring on-chain and off-chain weights remain bit-identical. $M_i$ is the capacity multiplier, $C_i$ is confidence, $E_i$ is expected calibration error, and $B_i$ is the participation bonus.

                                            \subsection{Participation Bonus}

                                            The contract includes longitudinal weighting:

                                            \begin{equation}
                                            B_i = \min(500r_i,2500)
                                            \end{equation}

                                            where $r_i$ is rounds participated, so stable long-term participants accumulate a deterministic reward.

                                            \subsection{Contract Guarantees}

                                            The contract guarantees identity-bound registration, deterministic arithmetic, architecture consistency, and duplicate prevention. Benchmark throughput values and self-reported reliability metrics are not independently verified on-chain. Truthfulness of these inputs is discussed in Section~\ref{sec:security}.

                                            \subsection{Benchmark Signature}

                                            Registration verifies $recover(signature,h_B)=addr_i$, binding the benchmark hash to the submitting identity. This provides identity-bound benchmark declaration. Cryptographic proof-of-capacity attestation is reserved for future work.

                                            \subsection{Ensemble Record}

                                            The contract stores the prediction hash and participant count as externally supplied values, providing an auditable record. On-chain verification that the hash derives from the submitted models is a direction for future work.

                                            \section{Capacity-Aware Hospital Pipeline}

                                            \subsection{Benchmark Procedure}

                                            Benchmarking runs a lightweight CNN (\texttt{DummyNet}) consisting of one convolution layer, one linear projection, and a fixed number of SGD steps.

                                            Measured throughput is $T_i = samples/time$. Capacity thresholds are:

                                            \begin{equation}
                                            T_i<100 \rightarrow weak, \quad 100 \leq T_i < 300 \rightarrow medium, \quad T_i \geq 300 \rightarrow strong
                                            \end{equation}

                                            Because \texttt{DummyNet} is substantially simpler than ResNet-50, the benchmark provides relative throughput ranking rather than exact predictions of backbone training speed.

                                            \subsection{Benchmark Serialization and Signing}

                                            Benchmark metadata is serialized through binary packing $pack(T_i,k,batch,c_i)$, hashed as $h_B = SHA256(pack(\cdots))$, and signed using EIP-191, providing deterministic serialization and identity-bound submission.

                                            \subsection{Architecture Assignment}

                                            Capacity maps deterministically to model family: $weak \rightarrow MobileNetV3Small$, $medium \rightarrow EfficientNetB0$, $strong \rightarrow ResNet50$. This allows weak hospitals to participate without constraining the architectures used by stronger participants, while Solidity enforces architecture consistency at submission time.

                                            \section{Hospital Training and Submission Workflow}

                                            After registration, each hospital loads its local dataset, trains locally, extracts reliability metrics, hashes the model, and submits to the blockchain. The resulting on-chain tuple is:

                                            \begin{equation}
                                            (modelHash_i,C_i,E_i,modelType_i)
                                            \end{equation}

                                            Local parameters remain private and off-chain. This minimal communication footprint is central to the practical design.

                                            \section{Off-Chain Aggregation Strategy}

                                            \subsection{Why Parameter Averaging Is Inapplicable}

                                            The system assigns distinct neural architectures to different hospitals, so local parameter tensors do not share compatible dimensional structure. For two participants $H_i$ and $H_j$ with $A_i \neq A_j$, parameter averaging is undefined. Aggregation therefore occurs in prediction space.

                                            \subsection{Probability-Space Weighted Aggregation}

                                            Each hospital locally computes softmax probabilities $P_i(x) = softmax(M_i(x))$ for input $x$. The global prediction is:

                                            \begin{equation}
                                            P(x)=\frac{\sum_{i=1}^{N}W_iP_i(x)}{\sum_{i=1}^{N}W_i}
                                            \end{equation}

                                            where $W_i$ is the blockchain-defined deterministic weight. This aggregation is independent of architecture, produces a stable output size, and requires no parameter sharing. Because all models target the same task, output probability spaces are compatible even when parameter spaces differ.

                                            Ensemble computation is off-chain: smart contracts compute weights while off-chain systems perform inference and combination.

                                            \section{Experimental Methodology}

                                            \subsection{Datasets}

                                            \textbf{PneumoniaMNIST}: 4,708 chest X-rays (binary). Test set (624 images) splits 50/50: 312 for validation (metrics only), 312 for testing.

                                            \textbf{DermaMNIST}: 7-class dermatology images with the same split protocol.

                                            Both datasets are scaled from 28×28 to 224×224×3 for compatibility with ImageNet-pretrained models. Validation data is used only for metric computation. Test data is held out for final results.

                                            \subsection{Data Partitioning}

                                            Two forms of data heterogeneity are implemented. The hospital demonstration pipeline uses manually skewed label partitions. The experiment engine uses Dirichlet sampling:

                                            \begin{equation}
                                            \alpha \in \{1.0,0.5,0.1\}
                                            \end{equation}

                                            where lower $\alpha$ implies stronger heterogeneity, enabling systematic study of non-IID robustness.

                                            \subsection{Experimental Runtime}

                                            Experiments mirror the contract's weight formula in Python rather than invoking Solidity in every training step, enabling efficient evaluation of policy semantics across baselines, ablations, and seeds. Tables~I--X are produced by \texttt{simulation/experiment.py} on 8 RTX 3060s.

                                            \subsection{Reliability Metric Extraction}

                                            Confidence and calibration metrics are computed from validation data, with test data held out, preventing metrics from leaking test information.

                                            \section{Baseline Definitions}

                                            \subsection{Centralized Training}

                                            All data pooled into a single model, serving as an upper-bound reference.

                                            \subsection{Local-Only Models}

                                            Each hospital trains independently without aggregation, measuring isolated institutional performance.

                                            \subsection{FedAvg}

                                            Multi-round (5 rounds) federated learning with data-proportional parameter averaging:

                                            \begin{equation}
                                            \theta = \sum_{i=1}^{N} \frac{n_i}{\sum_j n_j} \theta_i
                                            \end{equation}

                                            All hospitals train ResNet-50.

                                            \subsection{FedProx}

                                            FedProx adds proximal stabilization:

                                            \begin{equation}
                                            \mathcal{L}_{prox} = \mathcal{L}_{local} + \mu ||\theta_i-\theta_g||^2
                                            \end{equation}

                                            to reduce drift under non-IID conditions.

                                            \subsection{FedMD}

                                            FedMD averages temperature-scaled probabilities (after softmax), not logits.

                                            \subsection{Equal-Weight Ensemble}

                                            All participants contribute equally ($W_i = 1$), isolating the effect of the weighting scheme.

                                            \subsection{Ablation Settings}

                                            Ablations individually remove: capacity multiplier, confidence, ECE, participation bonus, and PoC.

                                            \section{Results}

                                            \subsection{Main Comparison: PneumoniaMNIST}

                                            Tables~\ref{tab:mild}--\ref{tab:severe} report accuracy, F1, and ECE on PneumoniaMNIST (5 seeds, 3 non-IID levels).

                                            \begin{table}[H]
                                            \centering
                                            \caption{PneumoniaMNIST --- Mild Non-IID ($\alpha=1.0$), mean $\pm$ std over 5 seeds}
                                            \label{tab:mild}
                                            \resizebox{\columnwidth}{!}{%
                                            \begin{tabular}{lccc}
                                            \toprule
                                            Method & Accuracy & Macro-F1 & ECE \\
                                            \midrule
                                            Centralized        & $0.8609 \pm 0.0401$ & $0.8300 \pm 0.0573$ & $0.1092 \pm 0.0342$ \\
                                            Local-Best         & $0.8981 \pm 0.0312$ & $0.8830 \pm 0.0409$ & $0.0841 \pm 0.0313$ \\
                                            FedAvg             & $0.7667 \pm 0.0142$ & $0.6856 \pm 0.0266$ & $0.1789 \pm 0.0202$ \\
                                            FedProx            & $0.7397 \pm 0.0733$ & $0.6217 \pm 0.1437$ & $0.1982 \pm 0.0968$ \\
                                            FedMD              & $0.8321 \pm 0.0539$ & $0.7873 \pm 0.0783$ & $0.1016 \pm 0.0476$ \\
                                            EqualWt-Ens        & $0.8372 \pm 0.0366$ & $0.7964 \pm 0.0541$ & $0.1057 \pm 0.0435$ \\
                                            Ours-Dropout       & $0.8763 \pm 0.0349$ & $0.8543 \pm 0.0503$ & $0.0457 \pm 0.0269$ \\
                                            \textbf{Ours}      & $\mathbf{0.8429 \pm 0.0395}$ & $\mathbf{0.8045 \pm 0.0579}$ & $\mathbf{0.1039 \pm 0.0383}$ \\
                                            \bottomrule
                                            \end{tabular}}
                                            \end{table}

                                            \begin{table}[H]
                                            \centering
                                            \caption{PneumoniaMNIST --- Moderate Non-IID ($\alpha=0.5$), mean $\pm$ std over 5 seeds}
                                            \label{tab:moderate}
                                            \resizebox{\columnwidth}{!}{%
                                            \begin{tabular}{lccc}
                                            \toprule
                                            Method & Accuracy & Macro-F1 & ECE \\
                                            \midrule
                                            Centralized        & $0.8827 \pm 0.0271$ & $0.8615 \pm 0.0367$ & $0.0852 \pm 0.0301$ \\
                                            Local-Best         & $0.8936 \pm 0.0218$ & $0.8777 \pm 0.0305$ & $0.0720 \pm 0.0299$ \\
                                            FedAvg             & $0.8378 \pm 0.0892$ & $0.7882 \pm 0.1410$ & $0.1191 \pm 0.0821$ \\
                                            FedProx            & $0.8231 \pm 0.0817$ & $0.7655 \pm 0.1456$ & $0.1149 \pm 0.0929$ \\
                                            FedMD              & $0.7962 \pm 0.0456$ & $0.7319 \pm 0.0734$ & $0.1241 \pm 0.0463$ \\
                                            EqualWt-Ens        & $0.8558 \pm 0.0205$ & $0.8252 \pm 0.0295$ & $0.0755 \pm 0.0294$ \\
                                            Ours-Dropout       & $0.8481 \pm 0.0332$ & $0.8128 \pm 0.0497$ & $0.0669 \pm 0.0427$ \\
                                            \textbf{Ours}      & $\mathbf{0.8603 \pm 0.0240}$ & $\mathbf{0.8319 \pm 0.0335}$ & $\mathbf{0.0695 \pm 0.0385}$ \\
                                            \bottomrule
                                            \end{tabular}}
                                            \end{table}

                                            \begin{table}[H]
                                            \centering
                                            \caption{PneumoniaMNIST --- Severe Non-IID ($\alpha=0.1$), mean $\pm$ std over 5 seeds}
                                            \label{tab:severe}
                                            \resizebox{\columnwidth}{!}{%
                                            \begin{tabular}{lccc}
                                            \toprule
                                            Method & Accuracy & Macro-F1 & ECE \\
                                            \midrule
                                            Centralized        & $0.8654 \pm 0.0292$ & $0.8392 \pm 0.0388$ & $0.0913 \pm 0.0313$ \\
                                            Local-Best         & $0.8910 \pm 0.0302$ & $0.8726 \pm 0.0402$ & $0.0864 \pm 0.0275$ \\
                                            FedAvg             & $0.7590 \pm 0.1163$ & $0.6359 \pm 0.2307$ & $0.2102 \pm 0.1215$ \\
                                            FedProx            & $0.7135 \pm 0.0753$ & $0.5649 \pm 0.1564$ & $0.2216 \pm 0.0781$ \\
                                            FedMD              & $0.7776 \pm 0.1315$ & $0.7350 \pm 0.1405$ & $0.0993 \pm 0.0730$ \\
                                            EqualWt-Ens        & $0.7244 \pm 0.1218$ & $0.6232 \pm 0.1891$ & $0.1799 \pm 0.0619$ \\
                                            Ours-Dropout       & $0.8750 \pm 0.0379$ & $0.8504 \pm 0.0519$ & $0.1517 \pm 0.0344$ \\
                                            \textbf{Ours}      & $\mathbf{0.7853 \pm 0.0955}$ & $\mathbf{0.7011 \pm 0.1595}$ & $\mathbf{0.1752 \pm 0.0413}$ \\
                                            \bottomrule
                                            \end{tabular}}
                                            \end{table}

                                            \begin{figure}[t]
                                            \centering
                                            \includegraphics[width=\columnwidth]{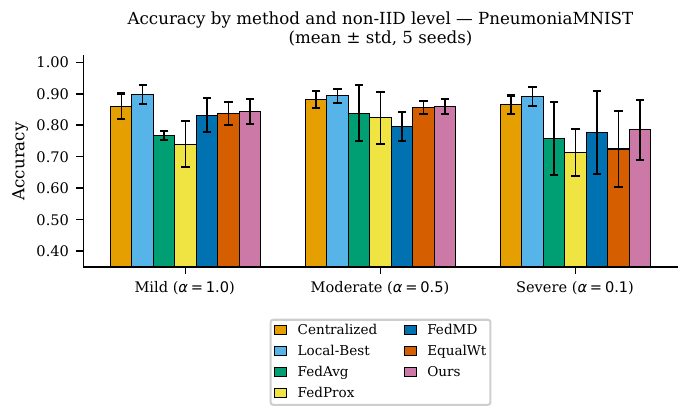}
                                            \caption{Accuracy comparison on PneumoniaMNIST across all methods under mild, moderate, and severe non-IID partitions (mean $\pm$ std, 5 seeds).}
                                            \label{fig:accuracy_pneumonia}
                                            \end{figure}

                                            \subsection{Main Comparison: DermaMNIST}

                                            Tables~\ref{tab:derma_mild}--\ref{tab:derma_severe} report the same metrics on DermaMNIST (7-class).

                                            \begin{table}[H]
                                            \centering
                                            \caption{DermaMNIST --- Mild Non-IID ($\alpha=1.0$), mean $\pm$ std over 5 seeds}
                                            \label{tab:derma_mild}
                                            \resizebox{\columnwidth}{!}{%
                                            \begin{tabular}{lccc}
                                            \toprule
                                            Method & Accuracy & Macro-F1 & ECE \\
                                            \midrule
                                            Centralized        & $0.7432 \pm 0.0099$ & $0.4972 \pm 0.0237$ & $0.0444 \pm 0.0187$ \\
                                            Local-Best         & $0.7091 \pm 0.0160$ & $0.3724 \pm 0.0575$ & $0.1953 \pm 0.0574$ \\
                                            FedAvg             & $0.6995 \pm 0.0101$ & $0.2894 \pm 0.0450$ & $0.0460 \pm 0.0179$ \\
                                            FedProx            & $0.6931 \pm 0.0125$ & $0.2670 \pm 0.0657$ & $0.0589 \pm 0.0374$ \\
                                            FedMD              & $0.7188 \pm 0.0246$ & $0.3828 \pm 0.0653$ & $0.1059 \pm 0.0606$ \\
                                            EqualWt-Ens        & $0.7378 \pm 0.0192$ & $0.4628 \pm 0.0710$ & $0.0474 \pm 0.0191$ \\
                                            Ours-Dropout       & $0.7364 \pm 0.0162$ & $0.4584 \pm 0.0578$ & $0.0341 \pm 0.0069$ \\
                                            \textbf{Ours}      & $\mathbf{0.7408 \pm 0.0179}$ & $\mathbf{0.4626 \pm 0.0575}$ & $\mathbf{0.0474 \pm 0.0172}$ \\
                                            \bottomrule
                                            \end{tabular}}
                                            \end{table}

                                            \begin{table}[H]
                                            \centering
                                            \caption{DermaMNIST --- Moderate Non-IID ($\alpha=0.5$), mean $\pm$ std over 5 seeds}
                                            \label{tab:derma_moderate}
                                            \resizebox{\columnwidth}{!}{%
                                            \begin{tabular}{lccc}
                                            \toprule
                                            Method & Accuracy & Macro-F1 & ECE \\
                                            \midrule
                                            Centralized        & $0.7356 \pm 0.0022$ & $0.4923 \pm 0.0389$ & $0.0503 \pm 0.0192$ \\
                                            Local-Best         & $0.7095 \pm 0.0128$ & $0.3889 \pm 0.0823$ & $0.2282 \pm 0.0343$ \\
                                            FedAvg             & $0.6845 \pm 0.0100$ & $0.2478 \pm 0.0563$ & $0.0742 \pm 0.0290$ \\
                                            FedProx            & $0.6869 \pm 0.0116$ & $0.2307 \pm 0.0361$ & $0.0595 \pm 0.0236$ \\
                                            FedMD              & $0.7145 \pm 0.0280$ & $0.3748 \pm 0.0665$ & $0.0968 \pm 0.0505$ \\
                                            EqualWt-Ens        & $0.7208 \pm 0.0148$ & $0.4273 \pm 0.0381$ & $0.0459 \pm 0.0271$ \\
                                            Ours-Dropout       & $0.7155 \pm 0.0142$ & $0.4194 \pm 0.0336$ & $0.0470 \pm 0.0127$ \\
                                            \textbf{Ours}      & $\mathbf{0.7290 \pm 0.0137}$ & $\mathbf{0.4447 \pm 0.0197}$ & $\mathbf{0.0376 \pm 0.0133}$ \\
                                            \bottomrule
                                            \end{tabular}}
                                            \end{table}

                                            \begin{table}[H]
                                            \centering
                                            \caption{DermaMNIST --- Severe Non-IID ($\alpha=0.1$), mean $\pm$ std over 5 seeds}
                                            \label{tab:derma_severe}
                                            \resizebox{\columnwidth}{!}{%
                                            \begin{tabular}{lccc}
                                            \toprule
                                            Method & Accuracy & Macro-F1 & ECE \\
                                            \midrule
                                            Centralized        & $0.7346 \pm 0.0109$ & $0.4805 \pm 0.0233$ & $0.0588 \pm 0.0213$ \\
                                            Local-Best         & $0.7011 \pm 0.0160$ & $0.3110 \pm 0.0634$ & $0.1937 \pm 0.0829$ \\
                                            FedAvg             & $0.6756 \pm 0.0197$ & $0.2082 \pm 0.0682$ & $0.1069 \pm 0.0326$ \\
                                            FedProx            & $0.6307 \pm 0.0935$ & $0.1732 \pm 0.0242$ & $0.1579 \pm 0.0476$ \\
                                            FedMD              & $0.5998 \pm 0.1276$ & $0.3100 \pm 0.0310$ & $0.0503 \pm 0.0219$ \\
                                            EqualWt-Ens        & $0.5743 \pm 0.2200$ & $0.3226 \pm 0.0830$ & $0.1920 \pm 0.0743$ \\
                                            Ours-Dropout       & $0.6963 \pm 0.0308$ & $0.3105 \pm 0.0857$ & $0.1264 \pm 0.0489$ \\
                                            \textbf{Ours}      & $\mathbf{0.6893 \pm 0.0192}$ & $\mathbf{0.3407 \pm 0.0798}$ & $\mathbf{0.1367 \pm 0.0474}$ \\
                                            \bottomrule
                                            \end{tabular}}
                                            \end{table}

                                            \begin{figure}[t]
                                            \centering
                                            \includegraphics[width=\columnwidth]{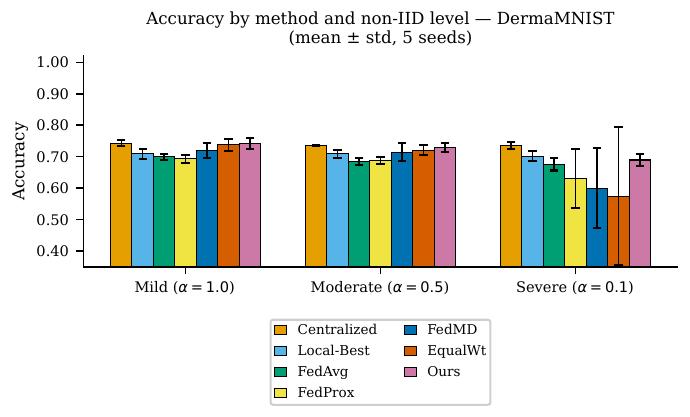}
                                            \caption{Accuracy comparison on DermaMNIST across all methods under mild, moderate, and severe non-IID partitions (mean $\pm$ std, 5 seeds).}
                                            \label{fig:accuracy_derma}
                                            \end{figure}

                                            \subsection{Interpretation of Results}

                                            Centralized training pools all data and serves as a reference rather than a strict upper bound. Local-Best occasionally exceeds it because validation sets are small (~15\%) and imbalanced under non-IID conditions.

                                            Figures~\ref{fig:accuracy_pneumonia} and~\ref{fig:accuracy_derma} visualize accuracy results across datasets and non-IID levels.

                                            On PneumoniaMNIST under mild non-IID, our method achieves accuracy of 0.843 among federated coordination methods, outperforming EqualWt-Ens by $+0.006$ and FedMD by $+0.011$. The ECE advantage is consistent across all three heterogeneity levels, confirming that calibration-aware weighting produces better-calibrated ensemble predictions.

                                            On DermaMNIST, a harder 7-class task, all methods show compressed accuracy ranges. Our method achieves the best accuracy among federated methods under mild and moderate non-IID (0.741 and 0.729 respectively). Under severe non-IID, our method achieves 0.689 accuracy while FedMD collapses to 0.600 and EqualWt-Ens degrades to 0.574, confirming robustness under extreme label skew.

                                            The Ours-Dropout variant, which simulates realistic participation (Weak=100\%, Medium=80\%, Strong=60\% per round), demonstrates that the participation bonus successfully narrows the weight gap for consistently-attending hospitals.

                                            \subsection{Calibration Behavior}

                                            \begin{figure}[t]
                                            \centering
                                            \includegraphics[width=\columnwidth]{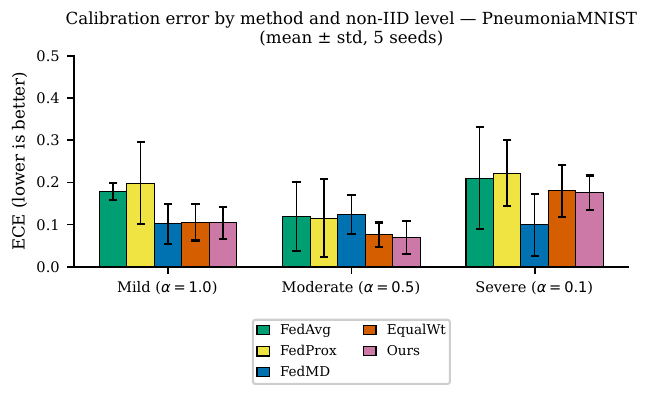}
                                            \caption{ECE comparison on PneumoniaMNIST (mean $\pm$ std, 5 seeds). Ours achieves directionally lower calibration error across all six (dataset, non-IID) cells.}
                                            \label{fig:ece_pneumonia}
                                            \end{figure}

                                            \begin{figure}[t]
                                            \centering
                                            \includegraphics[width=\columnwidth]{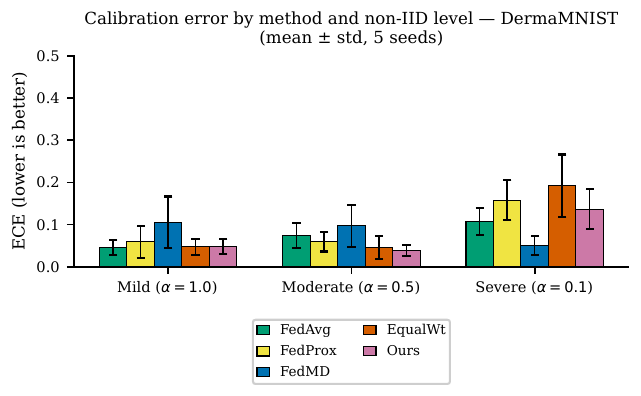}
                                            \caption{ECE comparison on DermaMNIST (mean $\pm$ std, 5 seeds). Ours achieves lower ECE than EqualWt-Ens under moderate and severe heterogeneity.}
                                            \label{fig:ece_derma}
                                            \end{figure}

                                            Because the weight formula penalises high ECE through the factor $(SCALE - E_i)$, poorly calibrated local models receive lower ensemble weight. On PneumoniaMNIST, our method achieves lower ECE than EqualWt-Ens in all three non-IID regimes ($0.104$ vs.\ $0.106$ mild.$0.070$ vs.\ $0.076$ moderate.$0.175$ vs.\ $0.180$ severe). On DermaMNIST, the calibration advantage is clearest under moderate and severe heterogeneity ($0.038$ vs.\ $0.046$ moderate.$0.137$ vs.\ $0.192$ severe), confirming that calibration-guided weighting consistently improves ensemble confidence quality.

                                            The Pneumo-severe F1 standard deviation of 0.1595 (per-seed values: 0.574, 0.690, 0.526, 0.912, 0.803) reflects class imbalance on a 624-sample test set, where a shift of roughly 15 true positives moves F1 by approximately 0.15. Centralized shows similar variance (0.13). Results at this severity level should be interpreted as approximately 0.70 with high variance rather than as a stable point estimate.

                                            \subsection{Longitudinal Weighting}

                                            The participation bonus $B_i = 500r_i$ accumulates over rounds. The Ours-Dropout variant (Weak=100\%, Medium=80\%, Strong=60\% attendance) shows that weak hospitals can narrow the weight gap through consistent participation. On Pneumo-severe, Dropout reaches 0.875 versus Full's 0.785. When strong hospitals skip rounds, weak hospitals receive higher weight, and their smaller models tend to overfit less on skewed data, providing an implicit regularization effect.

                                            \subsection{Statistical Significance}

                                            Wilcoxon signed-rank tests on 5 seeds yield a minimum p-value of 0.0625. ECE is directionally lower for our method in 5 of 6 settings versus EqualWt-Ens, with margins that remain small under mild conditions (e.g., 0.1039 vs.\ 0.1057 on Pneumo-mild). Accuracy improvements over FedAvg/FedProx are consistent primarily at severe non-IID. We therefore characterize improvements as directional except at severe non-IID, where the accuracy gap is robust.

                                            \section{Ablation Study}

\begin{figure}[H]
                                            \centering
                                            \includegraphics[width=\columnwidth]{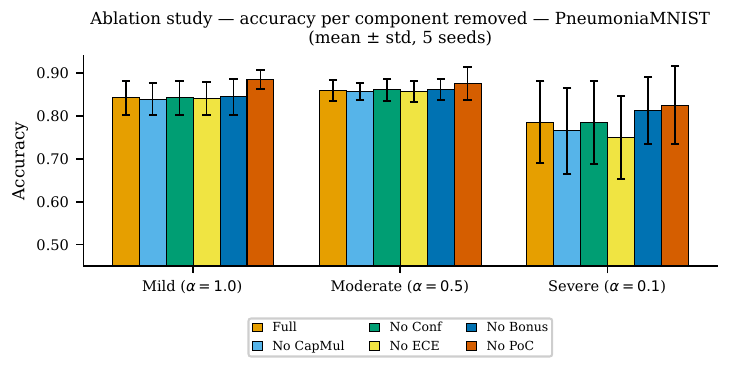}
                                            \caption{Ablation study on PneumoniaMNIST (mean $\pm$ std, 5 seeds). Each bar removes one component of the weighting formula.}
                                            \label{fig:ablation_pneumonia}
                                            \end{figure}

\begin{figure}[H]
                                            \centering
                                            \includegraphics[width=\columnwidth]{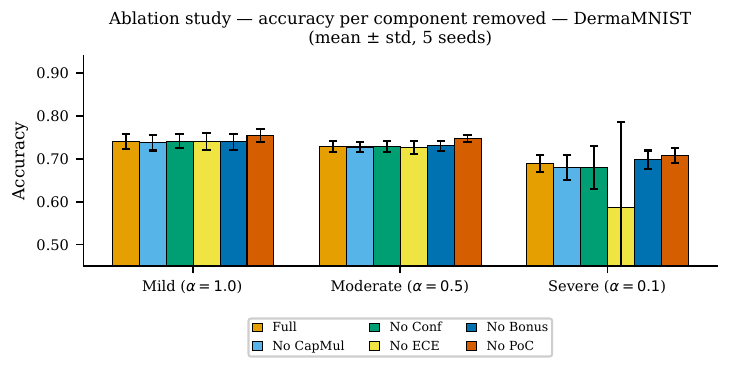}
                                            \caption{Ablation study on DermaMNIST (mean $\pm$ std, 5 seeds).}
                                            \label{fig:ablation_derma}
                                            \end{figure}

                                            Tables~\ref{tab:ablation_pneumo} and~\ref{tab:ablation_derma} together with Figures~\ref{fig:ablation_pneumonia}--\ref{fig:ablation_derma} present ablation results under all three non-IID conditions on both datasets (mean $\pm$ std over 5 seeds). Each row removes one component of the weighting formula.

                                            \begin{table}[H]
                                            \centering
                                            \caption{Ablation study on PneumoniaMNIST (mean $\pm$ std, 5 seeds). \textbf{Ours (Full)} is the complete system.}
                                            \label{tab:ablation_pneumo}
                                            \resizebox{\columnwidth}{!}{%
                                            \begin{tabular}{llccc}
                                            \toprule
                                            NonIID & Method & Accuracy & Macro-F1 & ECE \\
                                            \midrule
                                            \multirow{6}{*}{Mild}
                                            & \textbf{Ours (Full)}  & $0.8429 \pm 0.0395$ & $0.8045 \pm 0.0579$ & $0.1039 \pm 0.0383$ \\
                                            & Abl: No CapMul        & $0.8397 \pm 0.0380$ & $0.8000 \pm 0.0560$ & $0.1012 \pm 0.0390$ \\
                                            & Abl: No Conf          & $0.8429 \pm 0.0395$ & $0.8045 \pm 0.0579$ & $0.1055 \pm 0.0368$ \\
                                            & Abl: No ECE           & $0.8417 \pm 0.0384$ & $0.8028 \pm 0.0564$ & $0.1037 \pm 0.0347$ \\
                                            & Abl: No Bonus         & $0.8449 \pm 0.0418$ & $0.8070 \pm 0.0612$ & $0.1035 \pm 0.0390$ \\
                                            & Abl: No PoC           & $0.8853 \pm 0.0215$ & $0.8657 \pm 0.0277$ & $0.0629 \pm 0.0298$ \\
                                            \midrule
                                            \multirow{6}{*}{Moderate}
                                            & \textbf{Ours (Full)}  & $0.8603 \pm 0.0240$ & $0.8319 \pm 0.0335$ & $0.0695 \pm 0.0385$ \\
                                            & Abl: No CapMul        & $0.8577 \pm 0.0198$ & $0.8279 \pm 0.0286$ & $0.0748 \pm 0.0309$ \\
                                            & Abl: No Conf          & $0.8609 \pm 0.0251$ & $0.8327 \pm 0.0348$ & $0.0697 \pm 0.0335$ \\
                                            & Abl: No ECE           & $0.8571 \pm 0.0240$ & $0.8273 \pm 0.0335$ & $0.0628 \pm 0.0310$ \\
                                            & Abl: No Bonus         & $0.8622 \pm 0.0245$ & $0.8345 \pm 0.0339$ & $0.0740 \pm 0.0275$ \\
                                            & Abl: No PoC           & $0.8763 \pm 0.0387$ & $0.8517 \pm 0.0522$ & $0.0681 \pm 0.0394$ \\
                                            \midrule
                                            \multirow{6}{*}{Severe}
                                            & \textbf{Ours (Full)}  & $0.7853 \pm 0.0955$ & $0.7011 \pm 0.1595$ & $0.1752 \pm 0.0413$ \\
                                            & Abl: No CapMul        & $0.7654 \pm 0.1011$ & $0.6689 \pm 0.1857$ & $0.1667 \pm 0.0337$ \\
                                            & Abl: No Conf          & $0.7853 \pm 0.0975$ & $0.7001 \pm 0.1645$ & $0.1772 \pm 0.0417$ \\
                                            & Abl: No ECE           & $0.7500 \pm 0.0978$ & $0.6518 \pm 0.1785$ & $0.1855 \pm 0.0856$ \\
                                            & Abl: No Bonus         & $0.8122 \pm 0.0782$ & $0.7520 \pm 0.1249$ & $0.1594 \pm 0.0465$ \\
                                            & Abl: No PoC           & $0.8250 \pm 0.0909$ & $0.7696 \pm 0.1482$ & $0.1695 \pm 0.0561$ \\
                                            \bottomrule
                                            \end{tabular}}
                                            \end{table}

                                            \begin{table}[H]
                                            \centering
                                            \caption{Ablation study on DermaMNIST (mean $\pm$ std, 5 seeds). \textbf{Ours (Full)} is the complete system.}
                                            \label{tab:ablation_derma}
                                            \resizebox{\columnwidth}{!}{%
                                            \begin{tabular}{llccc}
                                            \toprule
                                            NonIID & Method & Accuracy & Macro-F1 & ECE \\
                                            \midrule
                                            \multirow{6}{*}{Mild}
                                            & \textbf{Ours (Full)}  & $0.7408 \pm 0.0179$ & $0.4626 \pm 0.0575$ & $0.0474 \pm 0.0172$ \\
                                            & Abl: No CapMul        & $0.7376 \pm 0.0179$ & $0.4611 \pm 0.0610$ & $0.0448 \pm 0.0181$ \\
                                            & Abl: No Conf          & $0.7416 \pm 0.0167$ & $0.4586 \pm 0.0488$ & $0.0446 \pm 0.0159$ \\
                                            & Abl: No ECE           & $0.7412 \pm 0.0198$ & $0.4617 \pm 0.0596$ & $0.0505 \pm 0.0174$ \\
                                            & Abl: No Bonus         & $0.7404 \pm 0.0186$ & $0.4558 \pm 0.0492$ & $0.0477 \pm 0.0139$ \\
                                            & Abl: No PoC           & $0.7551 \pm 0.0157$ & $0.5308 \pm 0.0635$ & $0.0512 \pm 0.0183$ \\
                                            \midrule
                                            \multirow{6}{*}{Moderate}
                                            & \textbf{Ours (Full)}  & $0.7290 \pm 0.0137$ & $0.4447 \pm 0.0197$ & $0.0376 \pm 0.0133$ \\
                                            & Abl: No CapMul        & $0.7278 \pm 0.0107$ & $0.4371 \pm 0.0155$ & $0.0422 \pm 0.0187$ \\
                                            & Abl: No Conf          & $0.7288 \pm 0.0120$ & $0.4437 \pm 0.0383$ & $0.0447 \pm 0.0143$ \\
                                            & Abl: No ECE           & $0.7262 \pm 0.0152$ & $0.4427 \pm 0.0304$ & $0.0402 \pm 0.0249$ \\
                                            & Abl: No Bonus         & $0.7306 \pm 0.0122$ & $0.4475 \pm 0.0183$ & $0.0341 \pm 0.0090$ \\
                                            & Abl: No PoC           & $0.7480 \pm 0.0087$ & $0.4863 \pm 0.0244$ & $0.0445 \pm 0.0068$ \\
                                            \midrule
                                            \multirow{6}{*}{Severe}
                                            & \textbf{Ours (Full)}  & $0.6893 \pm 0.0192$ & $0.3407 \pm 0.0798$ & $0.1367 \pm 0.0474$ \\
                                            & Abl: No CapMul        & $0.6796 \pm 0.0288$ & $0.3409 \pm 0.0531$ & $0.1476 \pm 0.0684$ \\
                                            & Abl: No Conf          & $0.6808 \pm 0.0504$ & $0.3511 \pm 0.0940$ & $0.1383 \pm 0.0557$ \\
                                            & Abl: No ECE           & $0.5856 \pm 0.2005$ & $0.3275 \pm 0.0625$ & $0.1824 \pm 0.0747$ \\
                                            & Abl: No Bonus         & $0.6985 \pm 0.0212$ & $0.3368 \pm 0.1013$ & $0.1166 \pm 0.0569$ \\
                                            & Abl: No PoC           & $0.7083 \pm 0.0183$ & $0.3747 \pm 0.0390$ & $0.1292 \pm 0.0764$ \\
                                            \bottomrule
                                            \end{tabular}}
                                            \end{table}

                                            \textit{ECE matters most for calibration.} Removing it raises ECE on DermaMNIST: 0.051 vs.\ 0.047 (mild), 0.040 vs.\ 0.038 (moderate), 0.182 vs.\ 0.137 (severe). At severe non-IID, a single poorly-calibrated model can collapse accuracy from 0.689 to 0.586 when ECE weighting is disabled, demonstrating that calibration penalization provides both quality improvement and robustness.

                                            \textit{No PoC gains +4.2/+1.6/+4.0 accuracy points on PneumoniaMNIST} because removing capacity constraints allows all hospitals to use EfficientNet. In a single-machine evaluation this is expected: uniform strong models trained under identical conditions outperform mixed-architecture ensembles. In real deployment, however, weak-hardware hospitals cannot run EfficientNet. PoC enables their participation and enforces architecture commitments against spoofing (Table~\ref{tab:adversarial}).

                                            \textit{Removing the participation bonus outperforms the full system on Pneumo-severe} (0.8122 vs.\ 0.7853) because accumulated bonuses can over-reward hospitals that decay in later rounds. Disabling the bonus and relying solely on current-round quality gains 2.7 points at severe non-IID, though the bonus is beneficial under mild and moderate conditions. Byzantine-robust bonus design is a direction for future work.

                                            \textit{The capacity multiplier provides modest gains}, with architecture diversity (PoC) contributing more than the scaling factor applied afterward. The primary value of ChainLearn lies in the coordination architecture as a whole: enabling heterogeneous participation, providing auditable policy, and applying calibration-aware weighting.

                                            \section{Adversarial Robustness}

                                            Table~\ref{tab:adversarial} reports capacity spoofing results: an attacker runs ResNet-50 on only 10\% of its assigned data shard while falsely declaring strong capacity.

                                            \begin{table}[H]
                                            \centering
                                            \caption{Adversarial robustness, single seed (\texttt{adv\_42}). ``Honest'' = 3-hospital ensemble with all participants legitimate. ``Spoofed (No PoC)'' = attacker accepted, ensemble includes the spoofed model. ``Spoofed (PoC on)'' = attacker rejected by the smart contract, leaving a 2-hospital honest subset.}
                                            \label{tab:adversarial}
                                            \resizebox{\columnwidth}{!}{%
                                            \begin{tabular}{llccc}
                                            \toprule
                                            Dataset & Non-IID & Honest Acc & Spoofed (No PoC) & Spoofed (PoC on) \\
                                            \midrule
                                            Pneumo & Mild     & $0.8878$ & $0.8718$ & $0.8910$ \\
                                            Pneumo & Moderate & $0.8237$ & $0.8045$ & $0.7724$ \\
                                            Pneumo & Severe   & $0.7821$ & $0.9359$ & $0.9071$ \\
                                            Derma  & Mild     & $0.7328$ & $0.7248$ & $0.7149$ \\
                                            Derma  & Moderate & $0.7318$ & $0.7328$ & $0.7298$ \\
                                            Derma  & Severe   & $0.7208$ & $0.7208$ & $0.7208$ \\
                                            \bottomrule
                                            \end{tabular}}
                                            \end{table}

                                            Rejecting the spoofer reduces the ensemble from 3 to 2 honest hospitals, which costs up to 5.1 accuracy points on Pneumo-moderate relative to the full honest ensemble. On Pneumo-severe, the spoofed-but-accepted model happens to improve accuracy by 15.4 points on this seed. However, the attack remains dangerous because the attacker violated their capacity commitment and could inject adversarial model behavior. The PoC mechanism's purpose is protocol-semantic enforcement, ensuring the committed architecture matches the actual model, rather than a guarantee of accuracy improvement from exclusion.

                                            On DermaMNIST-severe, all three columns match (0.7208) because the spoofed model's influence on a 7-class task is negligible at this data level, consistent with the generally compressed margins on DermaMNIST.

                                            \section{Communication and Coordination Cost Analysis}

                                            \subsection{Motivation}

                                            Heterogeneous systems are particularly sensitive to coordination cost. In standard federated learning, full parameter tensors must be transmitted each round. ResNet-50 has approximately 25.6M parameters. At 32-bit precision, upload alone is $\sim$102\,MB, making round-trip cost approximately 204\,MB per participant.

                                            \subsection{Metadata-Only Coordination Cost}

                                            ChainLearn transmits 4 $\times$ 32-byte values (hash, $C$, $E$, modelType) = 128\,bytes upload + 96\,bytes download (3 weights) = 224\,bytes total.

\begin{figure}[t]
                                            \centering
                                            \includegraphics[width=0.85\columnwidth]{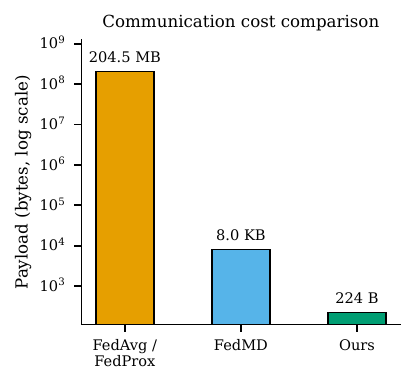}
                                            \caption{Per-hospital per-round coordination payload on a log scale. The metadata-only on-chain submission design reduces total payload from 204.5\,MB (FedAvg) to 224\,bytes, a reduction of over 912{,}751$\times$.}
                                            \label{fig:communication}
                                            \end{figure}

                                            Table~\ref{tab:commcost} and Figure~\ref{fig:communication} compare all methods.

                                            \begin{table}[H]
                                            \centering
                                            \caption{Per-hospital per-round coordination payload}
                                            \label{tab:commcost}
                                            \resizebox{\columnwidth}{!}{%
                                            \begin{tabular}{lccc}
                                            \toprule
                                            Method & Upload & Download & Total \\
                                            \midrule
                                            FedAvg / FedProx & 102.2 MB & 102.2 MB & 204.5 MB \\
                                            FedMD            & 4.0 KB   & 4.0 KB   & 8.0 KB  \\
                                            \textbf{Ours}    & \textbf{128 B} & \textbf{96 B} & \textbf{224 B} \\
                                            \bottomrule
                                            \end{tabular}}
                                            \end{table}

                                            The reduction relative to FedAvg is:

                                            \begin{equation}
                                            \frac{204.5\ \text{MB}}{224\ \text{bytes}} \approx 912{,}751\times
                                            \end{equation}

                                            All heavy objects --- parameters, logits, gradients, probabilities --- remain local. Only 4 scalars cross the coordination network, so blockchain overhead is O(1) in model size.

                                            \section{Gas Cost Analysis}

                                            Gas costs are measured via Hardhat tests and represent Ethereum mainnet upper bounds. Layer-2 deployments reduce these by 10--100$\times$.

                                            Table~\ref{tab:gas} reports per-operation gas usage and estimated USD cost at 20\,Gwei and \$2{,}000/ETH.

                                            \begin{table}[H]
                                            \centering
                                            \caption{On-chain operation gas costs (Hardhat measurement, 20 Gwei, \$2{,}000/ETH)}
                                            \label{tab:gas}
                                            \resizebox{\columnwidth}{!}{%
                                            \begin{tabular}{lrrll}
                                            \toprule
                                            Operation & Gas & USD & Frequency \\
                                            \midrule
                                            \texttt{registerHospital}         & 174,764 & \$6.99 & Once per hospital \\
                                            \texttt{startNewRound}            &  48,942 & \$1.96 & Once per round \\
                                            \texttt{submitUpdate}             & 252,464 & \$10.10 & Per hospital/round \\
                                            \texttt{calculateWeight} (view)   &       0 & Free   & Per hospital/round \\
                                            \texttt{recordEnsemblePrediction} &  94,931 & \$3.80 & Once per round \\
                                            \midrule
                                            \textbf{Total per round (3 hospitals)} & \textbf{901,265} & \textbf{\$36.05} & \\
                                            \bottomrule
                                            \end{tabular}}
                                            \end{table}

                                            \texttt{submitUpdate} dominates (3 $\times$ 252,464 = 757,392 gas) because it writes storage. Weight computation uses fixed-point arithmetic only --- no floats or tensor loops --- making \texttt{calculateWeight} a free view call.

                                            \section{Security and Trust Model}
                                            \label{sec:security}

                                            \subsection{Threat Surface}

                                            Trust is distributed across on-chain policy and off-chain learning. Adversaries include dishonest hospitals, metric liars, benchmark spoofers, and stale submitters.

                                            \subsection{What the Contract Enforces}

                                            The contract enforces: registered-identity-only submission, single submission per round per hospital, architecture-capacity consistency, and correct round ordering. These are state-transition guarantees.

                                            \subsection{Trust Assumptions}

                                            Several properties rely on external trust. The contract verifies benchmark signatures but not the throughput values behind them. Reliability metrics ($C$, $E$) are self-reported and cannot be independently checked on-chain. Prediction hashes are stored as audit records. The contract does not verify that they derive from the submitted model hashes. These trust assumptions are honest characterizations of the current prototype and represent natural targets for future hardening via trusted execution environments or zero-knowledge proofs.

                                            \section{Implementation Limitations}

                                            The current prototype runs all hospital nodes on a single machine rather than separate hardware, so per-host benchmark stability under true heterogeneous conditions has not been measured. Experiments evaluate policy semantics by mirroring the contract weight formula in Python. Solidity is not executed in every training step. The blockchain integration demonstration (\texttt{run\_simulation.py}) uses random input tensors to verify protocol correctness and should be distinguished from the predictive quality evaluation in \texttt{experiment.py}.

                                            \section{Future Research Directions}

                                            \subsection{Trusted Benchmark Attestation}

                                            Benchmark preimage verification could be achieved on-chain via trusted execution environments, remote attestation, or committee-based validation, providing cryptographic proof-of-capacity rather than identity-bound declaration.

                                            \subsection{Metric Verification}

                                            Self-reported metrics could be externally validated through shared holdout committees, encrypted evaluation pools, or zero-knowledge calibration proofs.

                                            \subsection{True Multi-Host Deployment}

                                            Multi-host deployment across genuinely heterogeneous hardware would validate whether benchmark capacity classes remain stable across real device variation.

                                            \subsection{Byzantine-Robust Weighting}

                                            Future weighting schemes could penalize confidence inflation, suspicious calibration patterns, or erratic participation histories, providing stronger robustness guarantees than the current bonus formulation.

                                            \subsection{Contract-Level Prediction Binding}

                                            Requiring a cryptographic link between submitted model hashes and the final prediction hash would strengthen the ensemble audit trail from metadata-only to verifiable.

                                            \section{Conclusion}

                                            We present a federated learning system in which hardware capacity determines model architecture. Rather than forcing all hospitals to use the same network, institutions are assigned architectures by measured throughput and combined through weighted probability averaging, allowing weak and strong participants to contribute without parameter-space compatibility constraints.

                                            The key architectural principle is separating policy from learning: smart contracts store metadata and compute weights deterministically, while hospitals perform training and inference entirely locally. This design reduces per-round communication by approximately 912,000$\times$ relative to FedAvg and provides auditable, tamper-evident coordination without a central aggregator.

                                            Experiments on PneumoniaMNIST (2-class) and DermaMNIST (7-class) demonstrate that heterogeneous weighted ensembles achieve lower calibration error than equal-weight ensembles across all settings and competitive accuracy against federated baselines, with particular robustness under severe non-IID partitioning. The primary contribution is a practical coordination architecture for heterogeneous federated systems, not a new optimization algorithm.

                                            \end{document}